\documentclass[lettersize,journal]{IEEEtran}
\usepackage{amsmath,amsfonts}
\usepackage{algorithmic}
\usepackage{algorithm}
\usepackage{array}
\usepackage[caption=false,font=normalsize,labelfont=sf,textfont=sf]{subfig}
\usepackage{textcomp}
\usepackage{stfloats}
\usepackage{url}
\usepackage{verbatim}
\usepackage{graphicx}
\usepackage{cite}
\usepackage{xcolor}
\usepackage{booktabs}
\usepackage{multirow}
\usepackage{makecell}
\usepackage{graphicx}
\begin{document}

\title{From Scene to Object: Text-Guided Dual-Gaze Prediction}

\author{Zehong Ke, Yanbo Jiang, Jinhao Li, Zhiyuan Liu, Yiqian Tu,\\
Qingwen Meng$^{\dagger}$, Heye Huang, Jianqiang Wang%
\thanks{
    This work is supported by the National Natural Science Foundation of China (GrantNo.52402502), and in part by the National Natural Science Foundation of China (No. 52221005). \textit{(Corresponding author: Qingwen Meng).}}%
\thanks{
    Zehong Ke, Yanbo Jiang, Jinhao Li, Zhiyuan Liu, Yiqian Tu, Qingwen Meng, and Jianqiang Wang are with the School of Vehicle and Mobility, Tsinghua University, Beijing 100084, China (e-mail: \{kzh21, jyb23, lijinhao24, liuzhiyu24, tuyq22\}@mails.tsinghua.edu.cn; \{mqwauto, wjqlws\}@tsinghua.edu.cn).}%
\thanks{
    Heye Huang is with Cho Chun Shik Graduate School of Mobility, Korea Advanced Institute of Science and Technology (KAIST), Daejeon 34141, Republic of Korea (e-mail: heye.huang@kaist.ac.kr).}}

\maketitle

\begin{abstract}
Interpretable driver attention prediction is crucial for human-like autonomous driving. However, existing datasets provide only scene-level global gaze rather than fine-grained object-level annotations, inherently failing to support text-grounded cognitive modeling. Consequently, while Vision-Language Models (VLMs) hold great potential for semantic reasoning, this critical data limitations leads to severe text-vision decoupling and visual-bias hallucinations. To break this bottleneck and achieve precise object-level attention prediction, this paper proposes a novel dual-branch gaze prediction framework, establishing a complete paradigm from data construction to model architecture. First, we construct G-W3DA, a object-level driver attention dataset. By integrating a multimodal large language model with the Segment Anything Model 3 (SAM3), we decouple macroscopic heatmaps into object-level masks under rigorous cross-validation, fundamentally eliminating annotation hallucinations. Building upon this high-quality data foundation, we propose the DualGaze-VLM architecture. This architecture extracts the hidden states of semantic queries and dynamically modulates visual features via a Condition-Aware SE-Gate, achieving intent-driven precise spatial anchoring. Extensive experiments on the W3DA benchmark demonstrate that DualGaze-VLM consistently surpasses existing state-of-the-art (SOTA) models in spatial alignment metrics, notably achieving up to a 17.8\% improvement in Similarity (SIM) under safety-critical scenarios. Furthermore, a human evaluation test reveals that the attention heatmaps generated by DualGaze-VLM are perceived as authentic by 88.22\% of human evaluators, proving its capability to generate rational cognitive priors.
\end{abstract}


\section{INTRODUCTION}

\IEEEPARstart{A}{s} autonomous driving technology progressively evolves toward an end-to-end paradigm, comprehensive and robust scene understanding has emerged as one of the most critical challenges. To enhance the interpretability of end-to-end models and achieve human-like driving behaviors, driver attention prediction has been widely studied to model how humans allocate visual focus. By learning the distribution of human gaze, autonomous systems can integrate visual saliency to prioritize safety-critical regions, effectively mimicking the ``Observation-Reasoning-Attention'' cognitive loop that characterizes expert human driving \cite{05-amadori2021hammerdrive}. This attention mechanism serves as a vital bridge, connecting high-level semantic reasoning with low-level spatial perception.

Despite the recognized importance of driver attention prediction, a critical issue hinders its advancement: existing scene-level gaze maps inherently fail to support text-grounded, object-level attention modeling. Typically, driver attention datasets---whether collected in simulators (e.g., DADA \cite{DADA-2000}, BDD-A \cite{BDDA}) or on real roads (e.g., DR(eye)VE \cite{DReyeVE})---are constructed by aggregating raw eye-tracking points, yielding a single, macroscopic global heatmap per scene. While these scene-level annotations successfully indicate where a driver generally looks, they lack explicit semantic labels and cognitive intent. This inherent ambiguity makes it difficult to elevate purely spatial saliency to the level of interpretable cognitive modeling.

The recent emergence of Vision-Language Models (VLMs) \cite{11-zhu2025llada, 13-renz2025simlingo} offers a promising direction for injecting semantic reasoning into driver attention prediction. By leveraging world knowledge acquired from massive image-text data, VLMs can generate both textual explanations and spatial heatmaps, thereby providing interpretable cognitive priors for autonomous driving systems. Pioneering works such as LLada \cite{W3DA} have demonstrated that VLM-based methods can jointly predict attention maps and causal reasoning texts. However, integrating general VLMs directly into this task exposes a fundamental modality gap. Since these methods rely solely on scene-level global heatmaps for visual supervision, they lack the fine-grained, object-level attention distributions that correspond to specific textual descriptions. Without explicit spatial-semantic references, the attention prediction and causal reasoning branches operate independently, directly leading to a severe text-vision decoupling phenomenon. A VLM might accurately reason in discrete text, producing a linguistically coherent description such as ``pay attention to the crossing pedestrian,'' yet its continuous attention prediction diffuses aimlessly across the scene, failing to visually ground the identified target.

Fundamentally, the absence of object-level annotations accounts for this disconnection. Existing paradigms merely present global heatmaps alongside raw images to prompt semantic explanations from a VLM, yet they fail to utilize these explanations to derive fine-grained object labels. Moreover, this simplistic annotation pipeline is highly susceptible to leakage from the raw images input, causing the model to hallucinate and describe objects that were never actually attended. This further exacerbates the text-vision misalignment and forces the model to default to pre-trained visual priors rather than genuinely grounded reasoning.

To overcome these fundamental limitations and break the decoupling bottleneck, this paper shifts the attention prediction paradigm from scene-level global estimation to text-guided object-level attention mapping. Recognizing that architectural innovation alone is insufficient without commensurate data support, we advocate a synergistic design encompassing both data construction and model architecture. At the data level, we introduce an automated pipeline that decouples holistic gaze maps into semantically grounded object masks, thereby establishing the missing link between linguistic descriptions and spatial attention. At the model level, we design a dual-branch network that leverages these fine-grained annotations to learn dynamic, text-driven attention generation.

The main contributions of this paper are summarized as follows:

\begin{itemize}
    \item \textbf{Automated Object-Level Annotation Pipeline:} We propose a novel, hallucination-free data annotation pipeline that fundamentally decouples coarse scene-level heatmaps into precise object-level attention masks. By intergrating VLM-based semantic parsing with cascaded SAM3 segmentation under a rigorous cross-validation mechanism, we construct G-W3DA—a large-scale dataset providing physically grounded supervision for text-vision cognitive alignment.
    
    \item \textbf{Query-Conditioned Dual-Branch Architecture:} We propose DualGaze-VLM, a dual-branch predictor designed to exploit the decoupled data paradigm. Unlike traditional models that output a single holistic heatmap, our architecture parallelly predicts both the macroscopic scene-level gaze and the microscopic object-level attention. By extracting semantic queries from reasoning context and employing a SE-Gate, it dynamically modulates visual features to generate target-specific gaze maps.
    
    \item \textbf{Empirical Validation of Object-Level Supervision:} Through extensive experiments and ablation studies, we demonstrate that object-level supervision elevates the precision and robustness of driver attention prediction. Our framework achieves SOTA performance on the W3DA benchmark, and a human evaluation test confirms that the generated heatmaps are perceived as authentic by 88.22\% of human evaluators, underscoring the cognitive plausibility of the learned priors.
\end{itemize}
The remainder of this paper is organized as follows. Section~II reviews related work on driver attention prediction and existing attention datasets. Section~III details the construction pipeline of the proposed G-W3DA dataset and the DualGaze-VLM architecture. Section~IV reports extensive experimental results, including quantitative comparisons with state-of-the-art methods, qualitative visualizations, ablation studies, and a human evaluation test. Finally, Section~V concludes the paper and outlines directions for future work.
\section{Related Work}
\subsection{Driver Attention Prediction}
Driver attention prediction aims to model how human drivers allocate visual attention during driving, and has evolved from early bottom-up visual saliency modeling, to task-driven prediction with multi-source priors, and more recently to VLM-enabled interpretable attention reasoning.
\subsubsection{Visual saliency and gaze estimation.} Early studies on driver attention prediction mainly followed a bottom-up visual saliency paradigm, which goal was to predict gaze distribution from low-level visual stimuli in the scene. To address the loss of spatial details in deep CNN during high-level semantic extraction, Cornia et al. proposed ML-Net \cite{01-cornia2016deep} to preserve fine-grained spatial information in saliency maps through multi-level feature fusion. Later research tended to adapt to generalized driving scenarios. Deng et al. \cite{02-deng2016does} introduced the road vanishing point as a static structural prior. Palazzi et al. proposed DR(eye)VE \cite{03-palazzi2018predicting} to further incorporate auxiliary inputs and used a multi-branch network to model dynamic visual attention, establishing an early benchmark for driver attention prediction. However, such methods are still based on pixel-level visual stimuli. They often fail to distinguish visually salient objects from task-relevant ones, which may lead to biased predictions in complex traffic scenes.
\subsubsection{Task-driven methods.} To overcome the limitations of visual saliency models in capturing driving tasks, later studies gradually shifted to a top-down task-driven paradigm by introducing driving states and external priors. For attention allocation in hazardous scenarios, Baee et al. \cite{04-baee2021medirl} proposed MEDIRL to inversely learn an implicit reward function from eye-tracking data. Amadori et al. \cite{05-amadori2021hammerdrive} further incorporated vehicle telemetry signals to gate the attention network, explicitly capturing the influence of driving actions on gaze distribution. Beyond driving states, external task priors have proven to be effective. Kotseruba et al. \cite{06-kotseruba2024scout+} introduced map information and navigation intent to guide the model toward key areas that are closely related to driving decisions. Chen et al. proposed FBLNet \cite{07-chen2023fblnet} to integrate historical state through a feedback-loop mechanism to simulate the accumulation of human driving experience. Zhou et al. proposed BKNet \cite{08-zhou2025behavior} and used knowledge embedding to connect traffic scenes with driving behaviors, enhancing task-level contextual understanding. Besides improving prediction accuracy, deployment efficiency and data dependency are also considered. Zhu et al. \cite{09-zhu2023unsupervised} explored unsupervised mining methods to reduce reliance on costly annotations. Zhao et al. \cite{10-zhao2025salm2} developed a lightweight SalM² model based on the Mamba architecture, significantly reducing computational cost while maintaining competitive performance. Although such methods made further progress in modeling task relevance, their reliance on additional information requires higher demands on data quality and availability.
\subsubsection{Interpretable prediction with VLMs.} The above methods have achieved notable progress, but still focus on fitting gaze distributions and lack a semantic understanding of why we look and what we look at. To address this limitation, recent studies have introduced VLMs to push driver attention prediction toward cognitive-level semantic reasoning. Zhou et al. proposed LLada \cite{11-zhu2025llada} and used a VLM to jointly generate attention heatmaps and causal reasoning text, enabling interpretability in attention prediction. Chen et al. \cite{12-chen2024gazexplain} and Renz et al. \cite{13-renz2025simlingo} further improved the understanding of the relationship between driving strategy and attention behavior by constructing a vision-language-action (VLA) loop. To address the limitations of VLMs in spatial perception and annotation cost, Wei et al. \cite{15-wei2025spatial} introduced LiDAR point clouds to enhance 3D geometric modeling, while Hamid et al. \cite{14-hamid2025fsdam} reduced the reliance on large-scale high-quality annotations through a few-shot learning strategy. Despite opening a new direction, such methods still face a mismatch in reasoning paradigms. Driver attention prediction requires fine-grained spatial modeling, whereas the core reasoning process of existing VLMs remains largely text-centered, making it difficult to translate textual reasoning into precise spatial gaze maps.
\subsection{Driver Attention Dataset}
The development of driver attention prediction relies heavily on the support of high-quality datasets. Existing data collection methods can be divided into simulator-based and naturalistic-driving-based using wearable eye trackers.
\subsubsection{Simulator-based datasets.} These datasets typically focus on hazard perception, abnormal event response, and visual saliency modeling in traffic scenes. For instance, DADA-2000 \cite{DADA-2000} emphasizes accident and near-accident scenarios and contains numerous driver gaze recordings collected during hazardous moments, providing a basis for hazard perception and risk-related attention research. BDD-A \cite{BDDA} is designed for attention analysis in critical driving situations, particularly focusing on changes in gaze distribution during abnormal events. TrafficSaliency \cite{CDNN} places more emphasis on visual saliency characteristics in traffic scenes and can be used to study the prediction of salient regions in driving environments. MAAD \cite{MAAD} further focuses on attention allocation and scene awareness during driving, extending the scope of datasets from “where to look” to “cognitive state.” Such datasets are built under controlled experimental conditions, which makes it easier to standardize the data collection process and highlight specific research questions.
\subsubsection{Datasets collected in real-world road environments.} This type of dataset places greater emphasis on authentic gaze behavior during natural driving. DR(eye)VE \cite{DReyeVE} is one of the representative datasets that provides videos in natural driving conditions together with corresponding driver gaze annotations, and has become a classic benchmark for driver gaze prediction. Gaze360 \cite{Gaze360} provides large-scale 3D gaze estimation data under full-view and free-head-motion settings, making it useful for improving model robustness to complex viewpoint changes and head pose variations. W3DA \cite{W3DA} prioritizes on 3D attention and gaze direction modeling, providing auxiliary support for 3D gaze reasoning. LBW \cite{LBW} serves as a complementary data source for gaze estimation in unconstrained outdoor environments, helping improve model generalization under less controlled conditions. Such datasets offer clear advantages in behavioral naturalness and task realism. However, they face challenges of high collection cost, noisy annotations and limited coverage of hazardous scenarios.
Overall, existing datasets mostly provide scene-level global gaze distributions and lack fine-grained object-level attention annotations. As a result, models can learn where attention is roughly directed in a scene, but struggle to establish explicit links between which object is attended to and why it is attended. In data construction, a common paradigm is to use VLM-based automatic annotation with human verification. However, when RGB images and global heatmaps are jointly involved in annotation or training process, models may assign attention to visually salient regions that are not actually highlighted in the heatmap. The lack of object-level supervision further aggravates the text-vision decoupling problem. Specifically, existing datasets encourage models to learn a fixed scene-level attention rather than capture the dynamic coupling among task, semantics and spatial targets. As a result, textual explanation and heatmap prediction often remain weakly connected during inference.

\section{Method}
\subsection{Overview and Problem Formulation}
In this study, we reformulate the driver attention prediction task from a traditional global vision regression problem into a multimodal, target-specific cognitive generation task. Formally, let $\mathcal{S}$ denote the set of front-view driving scenarios and $\mathcal{Q}$ denote the set of task-specific linguistic instructions. Given an input RGB frame $\mathbf{I}_{rgb} \in \mathbb{R}^{H \times W \times 3}$ and a specific query $\textbf{q} \in \mathcal{Q}$, the objective of our proposed model $\mathcal{F}_{\theta}$ is to construct a direct multimodal mapping from the visual-linguistic inputs to a unified cognitive output space:
\begin{equation}
    \mathcal{F}_{\theta} : \left( \mathbf{I}_{rgb}, \textbf{q} \right) \longmapsto \left( T, \mathbf{M}_{global}, \left\{ \mathbf{M}_{region}^{(k)} \right\}_{k=1}^{K} \right)
\end{equation}
where $T$ denotes the generated natural language reasoning sequence, $\mathbf{M}_{global} \in \mathbb{R}^{H \times W}$ represents the macroscopic global gaze probability map, and $\left\{ \mathbf{M}_{region}^{(k)} \right\}_{k=1}^{K} \subset \mathbb{R}^{H \times W}$ is a set of object-level fine-grained gaze maps corresponding to the $K$ explicitly identified objects parsed from $T$. For all predicted results, $\mathbf{M}(x,y) \in [0, 1]$ represents the normalized continuous visual saliency at pixel coordinates $(x,y)$.

To systematically achieve this, our methodology is driven by two core components. First, addressing the critical data bottleneck of coarse annotations, we design an automated object-level saliency dataset construction pipeline. By integrating Qwen3.5-Plus with the SAM model, we structurally decouple target-specific gaze masks from macroscopic heatmaps via a strict cross-validation mechanism. Second, to overcome the architectural misalignment between discrete linguistic tokens and continuous visual space, we propose DualGaze-VLM, a dual-branch multimodal network that extracts semantic region tokens to dynamically modulate visual features. Our key innovation is text-driven, object-conditioned saliency generation within a shared scene context. Instead of treating the entire visual field as a fixed saliency map, we explicitly condition the spatial decoder on semantic queries extracted from the VLM, enabling the shared visual backbone to produce distinct, task-specific gaze maps.

\begin{figure*}[t]
    \includegraphics[width=\textwidth]{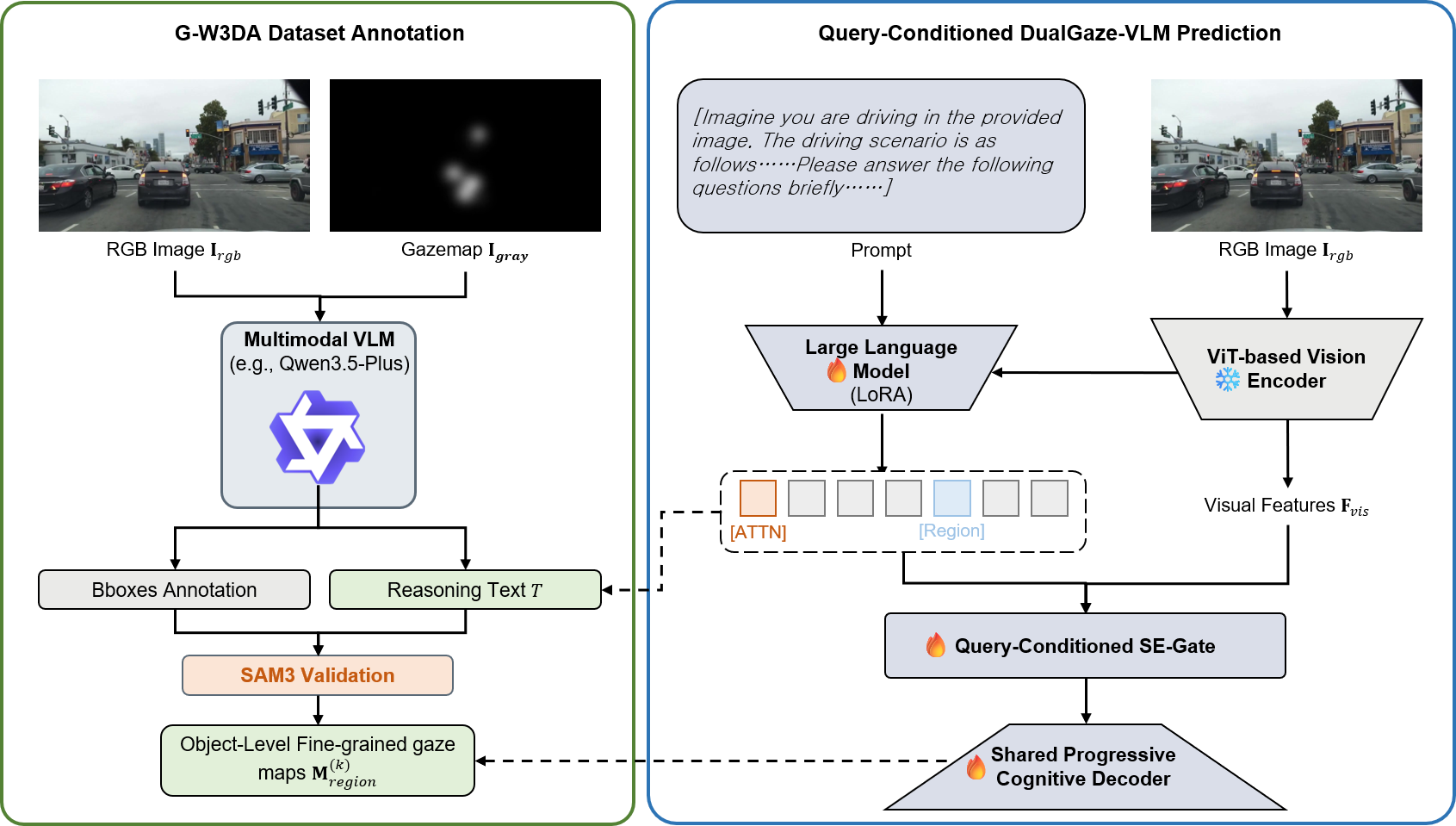}
    \caption{The overall architecture of the G-W3DA dataset generation pipeline and the proposed DualGaze-VLM framework. The left panel illustrates the decoupled spatial-semantic dataset annotation pipeline utilizing Qwen3.5-Plus and SAM3, showcasing the transformation from macroscopic heatmaps to object-level masks. The right panel details the query-conditioned network design, where semantic tokens extracted from the VLM dynamically modulate visual features via a Condition-Aware SE-Gate prior.}
    \label{fig:framework}
\end{figure*}

\subsection{Object-Level Gaze Dataset Construction Pipeline}
\label{s:pipeline}
A fundamental limitation of existing dataset annotation paradigms is ``Visual-Bias Hallucination''. In conventional dataset annotation paradigms, models are simultaneously presented with both a raw RGB image and a attention heatmap, and are then prompted to answer a series of scene-understanding questions. However, this joint-input approach lacks explicit mechanistic constraints to enforce strict compliance with the heatmap's guidance. Instead of faithfully interpreting the actual high-density areas indicated by the heatmap, the model tends to hallucinate and describe objects based on the raw image, even if those objects exhibit zero attention density. To avoid this hallucination, we propose an attention-faithful annotation pipeline driven by a novel dual-vision prompt strategy and rigorous cross-validation.

\begin{figure}[t]
    \centering
    \includegraphics[width=\columnwidth]{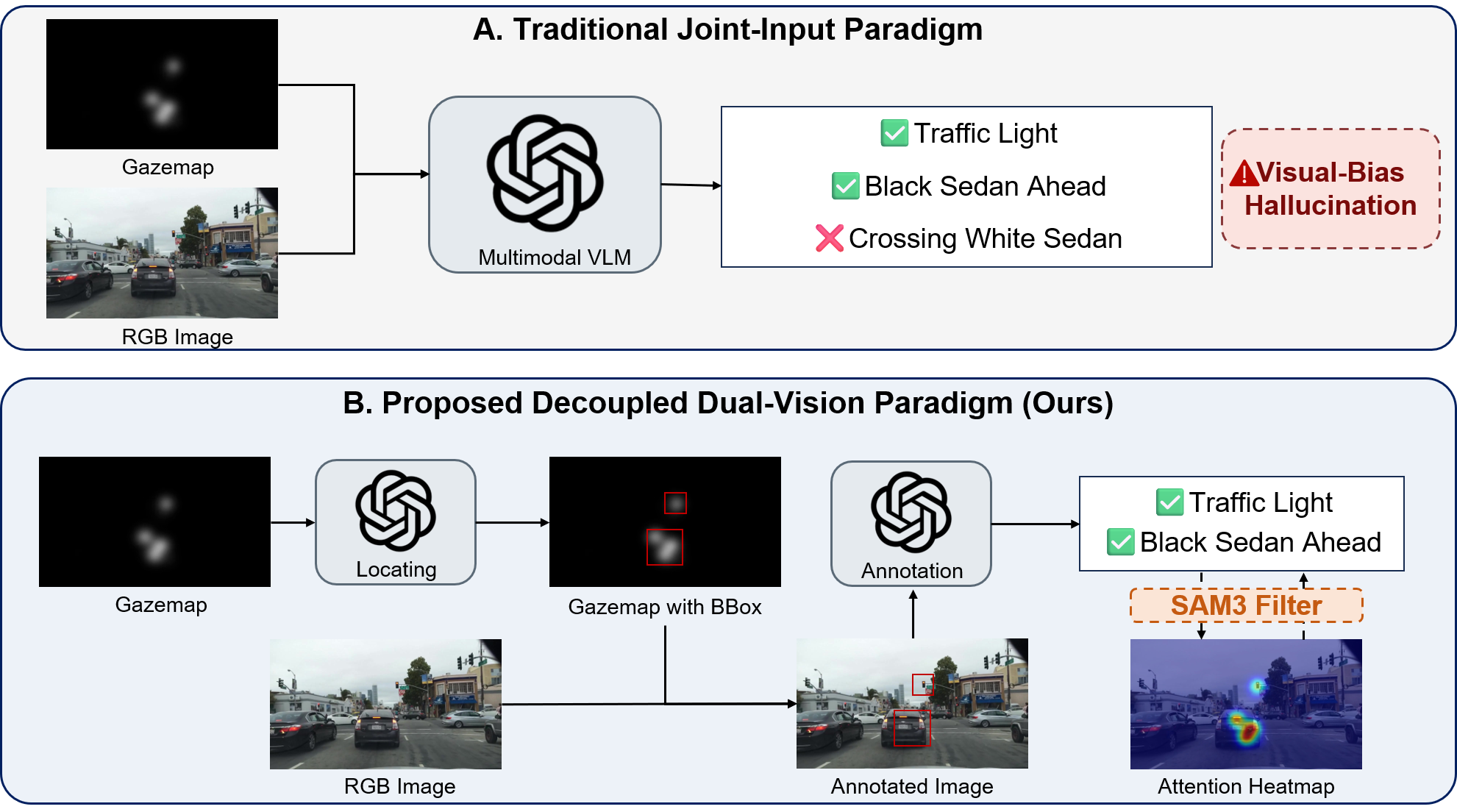}
    \caption{Comparison of dataset annotation paradigms. \textbf{Top (Traditional):} Feeding gaze maps and RGB images simultaneously often leads to visual-bias hallucination, where the VLM generates descriptions unaligned with the actual gaze distribution. \textbf{Bottom (Ours):} Our decoupled strategy restricts the initial spatial localization strictly to the grayscale heatmap, filtering task-irrelevant visual interference before semantic grounding.}
    \label{fig:pipeline_comparison}
\end{figure}

\subsubsection{Joint Spatial-Semantic Reasoning via Dual-Vision Prompting}
To enforce fidelity to the empirical human attention distribution while extracting rich semantics, we employ a dual-vision prompting strategy. The Qwen3.5-Plus receives two synchronized inputs simultaneously: the raw RGB driving scene $\mathbf{I}_{rgb}$ and its corresponding grayscale driver attention heatmap $\mathbf{I}_{gray}$ with higher pixel intensities representing a higher level of driver attention.

The generative model is instructed via a strict Chain-of-Thought (CoT) template to perform spatial-to-semantic mapping. Let $\mathcal{F}_{vlm}$ denote the inference process. The model first scans $\mathbf{I}_{gray}$ to identify salient raw spatial hotspots, yielding a set of bounding boxes $\mathbf{B}_{raw}$. Subsequently, it maps these coordinates back to $\mathbf{I}_{rgb}$ to semantically ground the objects, generating detailed descriptions. The VLM then removes duplicate instances, merging overlapping spatial boxes that belong to the identical semantic entity. The output is a structured JSON array containing $K$ unique regions:
\begin{equation}
    \mathcal{R} = \{ (\mathbf{B}_k, D_k, C_k) \}_{k=1}^K = \mathcal{F}_{vlm}(\mathbf{I}_{rgb}, \mathbf{I}_{gray}, P_{cot})
\end{equation}
where for the $k$-th target, $\mathbf{B}_k$ represents the merged bounding box $[x_{min}, y_{min}, x_{max}, y_{max}]$, $D_k$ serves as the semantic description of the attended object, and $C_k$ is the causal reasoning behind the driver's focus.

\subsubsection{Cascaded Zero-Shot Segmentation via SAM3}
While bounding boxes $\mathbf{B}_k$ provide robust coarse localization, pixel-level attention decoupling requires precise instance geometries. We employ SAM3, a unified foundation model for promptable segmentation, to ground the object via the generated descriptions $D_k$. However, raw linguistic descriptions generated by VLMs often suffer from semantic ambiguity, potentially yielding overly generalized object masks. 

To maximize object discovery while focusing on instances attended by human drivers, we design a cascaded fallback inference mechanism. For a given target description $D_k$, SAM3 predicts an initial set of candidate masks $\mathcal{S}_k = \{\mathbf{s}_{k,i}\}$ and corresponding confidence scores $\mathcal{C}_k = \{c_{k,i}\}$. In the primary pass, we establish a strict confidence threshold $\tau_{high}$. If $\max(\mathcal{C}_k) \ge \tau_{high}$, the high-confidence masks are retained. If the primary pass yields an empty set, indicating text-image misalignment or semantic ambiguity, the fallback strategy is triggered. The confidence threshold is relaxed to an ultra-low boundary $\tau_{low}$, and only the top-$N_{fallback}$ (where $N_{fallback} = 3$) masks are preserved based on their relative scores. Ultimately, a comprehensive set of candidate masks $\mathcal{S}_k$ is generated, serving as the foundational candidates for the subsequent cross-validation phase.

\subsubsection{Attention-Faithful Cross-Validation and Fusion}
\label{ss:SAM}
Geometric confidence scores derived from SAM3 do not necessarily equate to cognitive relevance. The remaining masks must accurately reflect the driver's actual visual focus to be retained. To rigorously filter out hallucinated objects or task-irrelevant background noise, we introduce a cross-validation protocol based on mean attention intensity, followed by intra-object mask fusion, as detailed in Algorithm \ref{alg:cross_val}.

\begin{algorithm}[h]
\caption{Attention-Faithful Cross-Validation and Intra-Object Fusion}
\label{alg:cross_val}
\begin{algorithmic}[1]
\REQUIRE Candidate Mask Set $\mathcal{S}_k$, Grayscale Heatmap $\mathbf{I}_{gray}^{*}$, Cognitive Threshold $\tau_{attn}=0.2$
\ENSURE Final Object-Level Mask $\mathbf{S}_{final}^{(k)}$
\STATE Initialize valid mask set $\mathcal{S}_{k}^{valid} \leftarrow \emptyset$
\FOR{\textbf{each} mask $\mathbf{s} \in \mathcal{S}_k$}
    \STATE Compute $\mu_{attn}(\mathbf{s})$ using Equation (\ref{eq:mu_attn})
\ENDFOR
\IF{$\exists \mathbf{s} \in \mathcal{S}_k \text{ such that } \mu_{attn}(\mathbf{s}) > \tau_{attn}$}
    \STATE $\mathcal{S}_{k}^{valid} \leftarrow \{ \mathbf{s} \in \mathcal{S}_k \mid \mu_{attn}(\mathbf{s}) > \tau_{attn} \}$
\ELSIF{$\max_{\mathbf{s} \in \mathcal{S}_k} \mu_{attn}(\mathbf{s}) > 0$}
    \STATE $\mathcal{S}_{k}^{valid} \leftarrow \{ \arg\max_{\mathbf{s} \in \mathcal{S}_k} \mu_{attn}(\mathbf{s}) \}$
\ELSE
    \STATE $\mathcal{S}_{k}^{valid} \leftarrow \{ \mathbf{0}^{H \times W} \}$ (Complete hallucination detected)
\ENDIF
\STATE $\mathbf{S}_{final}^{(k)} \leftarrow \bigvee_{\mathbf{s} \in \mathcal{S}_{k}^{valid}} \mathbf{s}$
\RETURN $\mathbf{S}_{final}^{(k)}$
\end{algorithmic}
\end{algorithm}

For each candidate mask $\mathbf{s} \in \{0,1\}^{H \times W}$, we compute its mean attention intensity $\mu_{attn}(\mathbf{s})$ by performing a discrete 2D integration over the normalized grayscale heatmap $\mathbf{I}_{gray}^{*}$:
\begin{equation}
    \mu_{attn}(\mathbf{s}) = \frac{\sum_{x=1}^{W}\sum_{y=1}^{H} \left(\mathbf{I}_{gray}^{*}(x,y) \cdot \mathbf{s}(x,y)\right)}{\sum_{x=1}^{W}\sum_{y=1}^{H} \mathbf{s}(x,y)}
    \label{eq:mu_attn}
\end{equation}

To systematically eliminate task-irrelevant proposals, the cross-validation protocol enforces a predefined cognitive threshold $\tau_{attn}$. Let $\mu_{max} = \max_{\mathbf{s} \in \mathcal{S}_k} \mu_{attn}(\mathbf{s})$ denotes the maximum attention intensity among all candidates in $\mathcal{S}_k$. The validated mask set $\mathcal{S}_{k}^{valid}$ is then derived via a hierarchical selection logic:
\begin{equation}
    \resizebox{0.91\columnwidth}{!}{$
    \mathcal{S}_{k}^{valid} := 
    \begin{cases} 
      \{ \mathbf{s} \in \mathcal{S}_k \mid \mu_{attn}(\mathbf{s}) > \tau_{attn} \}, & \text{if } \mu_{max} > \tau_{attn} \\
      \{ \arg\max_{\mathbf{s} \in \mathcal{S}_k} \mu_{attn}(\mathbf{s}) \}, & \text{if } 0 < \mu_{max} \le \tau_{attn} \\
      \{ \mathbf{0}^{H \times W} \}, & \text{if } \mu_{max} = 0
    \end{cases}
    $}
    \label{eq:hierarchical_logic}
\end{equation}
Critically, a zero maximum attention intensity indicates that the VLM completely hallucinated an object the driver did not focus on. The algorithm replaces this candidate with an all-zero tensor $\mathbf{0}^{H \times W}$, preserving the purity of the cognitive dataset. Finally, to generate a comprehensive representation for the semantic target, we perform intra-object mask fusion. All validated masks within $\mathcal{S}_{k}^{valid}$ are aggregated using a pixel-wise logical OR operation ($\bigvee$), yielding the final pure object mask $\mathbf{S}_{final}^{(k)}$. 

\subsubsection{Target Gaze Decoupling and Counterfactual Generation}
In the final phase, the rigorously verified fused spatial mask $\mathbf{S}_{final}^{(k)}$ is employed to fundamentally decouple the global gaze map. By applying a pixel-wise Hadamard product, we isolate the target-specific ground truth:
\begin{equation}
    \mathbf{M}_{region}^{(k)}(x,y) = \mathbf{M}_{global}(x,y) \odot \mathbf{S}_{final}^{(k)}(x,y)
\end{equation}

Through the execution of this rigorous Semantic-to-Spatial Validation Pipeline, we ultimately construct the Grounded-W3DA (G-W3DA) dataset. This dataset provides a robust, fine-grained data foundation for VLMs to master precise text-object cognitive grounding in complex physical scenes. While traditional driving datasets enforce a deterministic and static projection between driving scenes $\mathbf{I}_{rgb}$ and holistic gaze maps $\mathbf{M}_{global}$, our G-W3DA dataset extracts multiple distinct $\mathbf{M}_{region}^{(k)}$ from a single scene based on different semantic reasoning texts $D_k$. Consequently, by grounding fine-grained, object-level labels, we empower the VLMs to process the identical input image $\mathbf{I}_{rgb}$ alongside different task queries, yielding a task-specific spatial mask $\mathbf{M}_{task} \in \{\mathbf{M}_{global}, \mathbf{M}_{region}^{(k)}\}$.

\subsection{Query-Conditioned DualGaze-VLM Architecture}
To fully exploit the fine-grained counterfactual data generated by our pipeline, we propose DualGaze-VLM, an innovative multimodal cognitive architecture. Conventional driver attention models typically regress a single global gaze map, where linguistic reasoning and spatial prediction operate as loosely coupled branches lacking explicit mutual constraints. Consequently, such models may generate plausible text while failing to visually ground the specific objects. To bridge this semantic-spatial gap, DualGaze-VLM introduces a natural cross-modal regularization by predicting both the macroscopic global gaze and the text-guided microscopic object gaze. This dual-target strategy inherently forces the network to maintain strict alignment between its abstract cognitive reasoning and its continuous visual attention.
\subsubsection{Multimodal Visual Encoding}
Given the input RGB image $\mathbf{I}_{rgb} \in \mathbb{R}^{H \times W \times 3}$, the frozen vision encoder maps the non-overlapping patches into a higher-dimensional embedding space. The embeddings are processed through consecutive Vision Transformer (ViT) blocks to extract multi-scale visual features $\mathbf{F}_{vis} \in \mathbb{R}^{C \times H' \times W'}$, where $C$ is the channel dimension, and $H'$ and $W'$ represent the downsampled spatial dimensions. To preserve fine-grained visual details across varying semantic levels, intermediate multi-level features from the ViT are projected by dedicated mergers and directly added as residuals to the corresponding hidden states.

\subsubsection{Unified Semantic Query Extraction}
Whether predicting the global driving gaze or a specific object-level mask, the model must dynamically alter its spatial perception based on the task instruction. Due to the causal, autoregressive nature of VLMs, the generation of a textual sequence is strictly conditioned on its preceding representations. Consequently, the hidden state vector of the initiating token of a specific description serves as the direct generative basis, guiding the model's subsequent cognitive reasoning. As a predictive anchor, this leading token encodes the semantic intent of the following texts.

Based on this foundational principle, we formalize a unified extraction strategy tailored to our dual-target architecture. For the macroscopic global branch, we explicitly insert a special \texttt{[ATTN]} token at the beginning of the overall reasoning response, extracting its corresponding hidden state as the global semantic anchor $\mathbf{h}_{global} \in \mathbb{R}^C$. For the microscopic region branch, the model directly extracts the hidden state of the first token initiating the specific regional description to serve as the object-specific semantic anchor $\mathbf{h}_{region}^{(k)} \in \mathbb{R}^C$ for the $k$-th object. These extracted vectors, generalized as the unified semantic query $\mathbf{h}_{task} \in \mathbb{R}^C$, provide semantic queries for the subsequent visual feature modulation.

\subsubsection{Query-Conditioned SE-Gate Modulation}
Let $\mathbf{h}_{task} \in \mathbb{R}^C$ denote the semantic anchor extracted from the leading token. To translate this abstract cognitive intent into continuous spatial perception, we first employ a multi-head cross-attention (MHCA) mechanism. By treating the semantic anchor as the query and the visual features as the key and value pairs, the model dynamically extracts target-specific spatial features. The flattened visual feature map $\mathbf{F}'_{vis} \in \mathbb{R}^{(H'W') \times C}$ is projected into the attention space using learnable weight matrices $\mathbf{W}_Q, \mathbf{W}_K$, and $\mathbf{W}_V$:
\begin{equation}
\begin{aligned}
    \mathbf{Q} &= \mathbf{h}_{task}\mathbf{W}_Q \\
    \mathbf{K} &= \mathbf{F}'_{vis}\mathbf{W}_K \\
    \mathbf{V} &= \mathbf{F}'_{vis}\mathbf{W}_V \\
    \mathbf{F}_{attn} &= \text{Softmax}\left(\frac{\mathbf{Q} \mathbf{K}^\top}{\sqrt{d_k}}\right) \mathbf{V}
\end{aligned}
\end{equation}
where $d_k$ is the scaling factor. The output $\mathbf{F}_{attn} \in \mathbb{R}^{1 \times C}$ represents a task-specific semantic query. To preserve the original spatial geometries while injecting these localized semantic features, the reshaped spatial tensor $\mathbf{F}_{attn}$ is broadcast and merged with the original visual features via a residual connection $\mathbf{F}_{fused} = \mathbf{F}_{attn} \oplus \mathbf{F}_{vis}$.

In practice, naturalistic driving datasets inherently exhibit strong spatial priors, such as a severe center-bias. The network tends to rely heavily on the spatial layout of the visual features ($\mathbf{F}_{vis}$) instead of the semantic input $\mathbf{h}_{task}$. As a result, the generated gaze maps may be mismatched with the textual descriptions. To realize semantic-guided feature modulation, we compute channel weights via a squeeze-condition-excite pipeline \cite{SE-Block}. First, a global spatial summary $\mathbf{z} \in \mathbb{R}^C$ is squeezed via Global Average Pooling (GAP) across the spatial dimensions of $\mathbf{F}_{fused}$. To inject the semantic prior, this visual summary $\mathbf{z}$ is concatenated with the task-specific query $\mathbf{h}_{task} \in \mathbb{R}^C$. A dynamic channel-weighting vector $\mathbf{g} \in \mathbb{R}^C$ is then obtained through a bottleneck Multi-Layer Perceptron:
\begin{equation}
    \mathbf{g} = \sigma\left(\mathbf{W}_2 \cdot \delta(\mathbf{W}_1 [\mathbf{z} \parallel \mathbf{h}_{task}])\right)
\end{equation}
where $\parallel$ denotes feature concatenation. Functionally, this block acts as an information bottleneck where $\mathbf{W}_1 \in \mathbb{R}^{\frac{C}{r} \times 2C}$ compresses the multi-modal input governed by the reduction ratio $r$, and $\mathbf{W}_2 \in \mathbb{R}^{C \times \frac{C}{r}}$ restores the dimension. Through this paradigm, the network effectively isolates the most essential channel-wise dependencies. Finally, the Sigmoid function $\sigma(\cdot)$ maps these refined features into the range $(0, 1)$, yielding the dynamic channel attention coefficients $\mathbf{g}$.

Finally, the modulated visual feature is obtained via spatial broadcasting and element-wise multiplication:
\begin{equation}
    \mathbf{F}_{mod} = \mathbf{F}_{fused} \otimes \mathbf{g}
\end{equation}

The SE-Gate explicitly suppresses irrelevant background channels and excites target-specific features. This ensures that the identical visual backbone can dynamically modulate either global or region-specific representations for the subsequent decoder conditioned on different queries.

\subsubsection{Shared Progressive Cognitive Decoder}
To translate the modulated features into a continuous spatial heatmap, both the global and regional branches process their representations through a shared progressive cognitive decoder. To ensure training stability, the modulated features $\mathbf{F}_{mod}$ first undergo layer normalization (LN) across the channel dimension, yielding the normalized features $\mathbf{F}_{norm}$:
\begin{equation}
    \mathbf{F}_{norm} = \text{LN}(\mathbf{F}_{mod}) = \frac{\mathbf{F}_{mod} - \mu}{\sqrt{\sigma^2 + \epsilon}} \cdot \gamma + \beta
\end{equation}
where $\mu$ and $\sigma^2$ denote the channel-wise mean and variance of $\mathbf{F}_{mod}$. By computing statistics independently for each sample, LN guarantees highly stable gradient propagation under the limitation of small batch sizes.

\begin{figure}
    \centering
    \includegraphics[width=\linewidth]{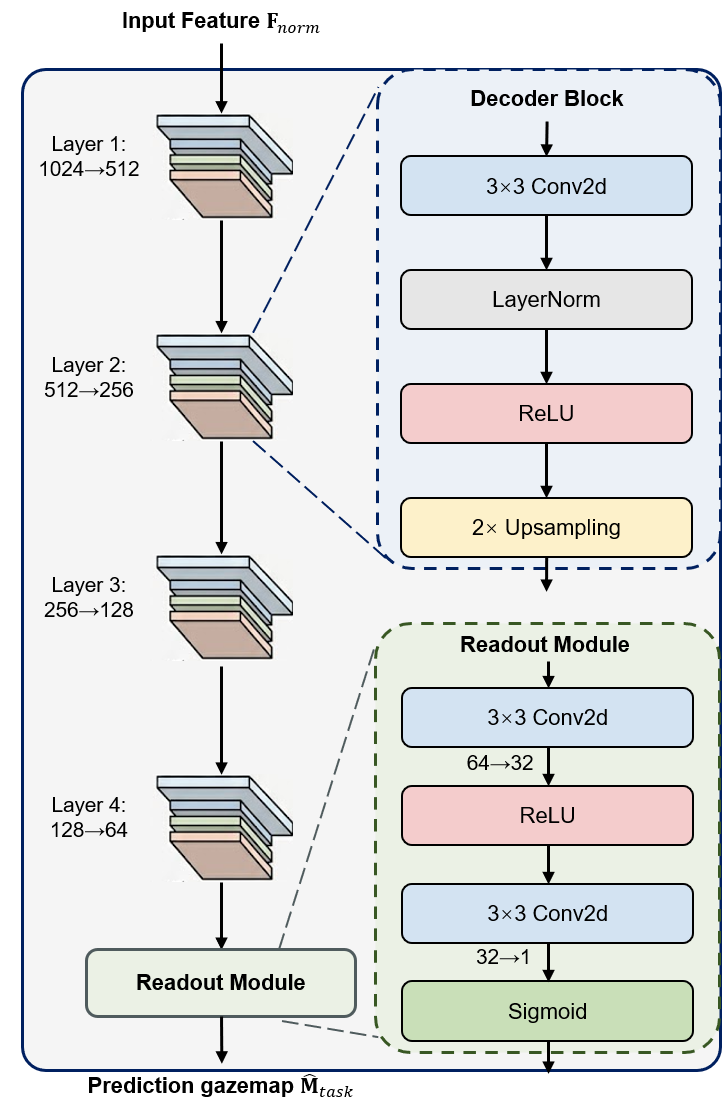} 
    \caption{Detailed architecture of the Shared Progressive Cognitive Decoder. A cascade of convolutional and upsampling blocks progressively reconstructs the spatial resolution, followed by a readout module to obtain a standardized $256 \times 256$ continuous gaze probability map $\hat{\mathbf{M}}_{task}$ tailored to the input cognitive query.}
    \label{fig:decoder_arch}
\end{figure}

Taking these stabilized features as input, the decoder explicitly follows a bottleneck design to generate a spatial gaze map. As illustrated in Fig. \ref{fig:decoder_arch}, the normalized features initially undergo convolutional downsampling to expand the receptive field. Subsequently, the decoder utilizes a cascade of convolutional blocks with upsampling layers to progressively restore the spatial resolution. Finally, a $1 \times 1$ convolutional layer with a Sigmoid activation yields the continuous probability map $\hat{\mathbf{M}}_{task} \in [0, 1]^{H_{out} \times W_{out}}$. Regardless of the input aspect ratio, the final spatial output is standardized to $H_{out} = W_{out} = 256$ in our work.
\subsection{Joint Optimization and Mixed Loss Strategy}
Our framework optimizes the unified query-conditioned predictions through a mixed loss objective. This function supervises the model across three distinct dimensions: linguistic causality, distribution alignment, and numerical fitting.

\subsubsection{Causal Language Modeling Loss}
The VLM backbone is optimized to generate accurate reasoning text using a standard auto-regressive cross-entropy loss over the generated tokens:
\begin{equation}
    \mathcal{L}_{text} = -\sum_{t=1}^{T} \log P(y_t | y_{<t}, \mathbf{I}_{rgb})
\end{equation}
where $\mathbf{I}_{rgb}$ is the input driving scene, $T$ denotes the total number of tokens, and $y_t$ represents the $t$-th generated token conditioned on the preceding context $y_{<t}$. The target sequence $\{y_t\}_{t=1}^T$ represents the ground truth linguistic annotations, specifically encompassing the object identifiers $C_k$ and their corresponding semantic reasoning descriptions $D_k$. By strictly supervising the generation of $C_k$ and $D_k$, we ensure that the subsequently extracted query vector $\mathbf{h}_{task}$ carries a valid cognitive prior for the spatial decoder.

\subsubsection{Spatial Distribution Alignment via KL Divergence}
For visual attention modeling, predicting the exact pixel-wise intensity is often less critical than capturing the topological distribution of human gaze. Therefore, we formalize the spatial prediction as a 2D probability distribution fitting problem. Inspired by \cite{W3DA}, we employ Kullback-Leibler (KL) Divergence to align the predicted spatial density with the ground truth. For a target gaze map $\mathbf{G}$ and prediction $\hat{\mathbf{M}}$, the loss is defined as:
\begin{equation}
    \mathcal{L}_{kld}(\mathbf{G}, \hat{\mathbf{M}}) = \sum_{x,y} \mathbf{G}(x,y) \log \left( \frac{\mathbf{G}(x,y)}{\hat{\mathbf{M}}(x,y) + \epsilon} \right)
\end{equation}
where $\epsilon$ ensures numerical stability, and both maps are normalized such that their sums equal 1. 

For the region branch, this topological alignment serves as the exclusive optimization objective, computed across all $K$ object-level gaze maps:
\begin{equation}
    \mathcal{L}_{region} = \sum_{k=1}^{K} \mathcal{L}_{kld}(\mathbf{G}_{region}^{(k)}, \hat{\mathbf{M}}_{region}^{(k)})
\end{equation}
For highly localized object-level attention, the objective is fundamentally to learn the relative spatial distribution rather than absolute numerical intensities. Using KL divergence as supervisor forces the probability density center of the predicted heatmap to faithfully mirror the true fixation spread, effectively capturing the topological distribution.

\subsubsection{Spatial-Weighted BCE for numerical fitting}
\begin{figure}[t]
    \centering
    \includegraphics[width=\linewidth]{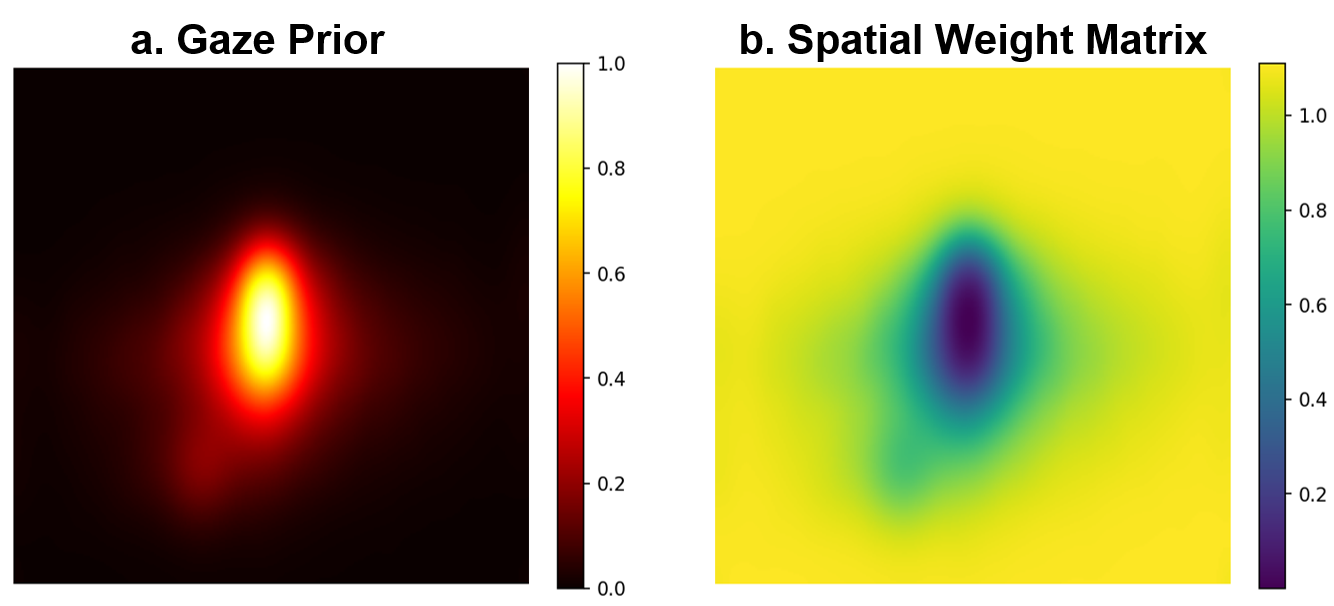} 
    \caption{Visualization of the strong center-bias prior via aggregated ground-truth statistics across the entire training dataset. Brighter and warmer colors indicate higher spatial frequencies of gaze points, revealing a severe concentration of attention around the road center straight ahead.}
    \label{fig:center_bias}
\end{figure}

While KL Divergence effectively ensures topological alignment, the global branch still demands fine-grained pixel-level numerical fitting across the entire visual field. However, as shown in Fig. \ref{fig:center_bias}, human driving gaze suffers from severe center bias toward straight ahead near the road's vanishing point. A standard Binary Cross-Entropy (BCE) loss would lazily overfit to this central region, critically sacrificing marginal prediction accuracy.
To mitigate this distribution imbalance, we propose a Spatial-Weighted BCE loss:
\begin{equation}
    \mathcal{L}_{wbce} = \frac{1}{N} \sum_{x,y} \mathbf{W}_{spatial}(x,y) \cdot \ell_{bce}(x,y)
\end{equation}
where $N = H \times W$, $\ell_{bce}(x,y)$ represents the standard Binary Cross-Entropy at the spatial coordinate $(x,y)$.
 $\mathbf{W}_{spatial}$ is a pre-computed prior matrix, serving as the core of our proposed mechanism. Specifically, let $\tilde{F}(x,y) \in [0, 1]$ denote the normalized spatial frequency as illustrated in Fig. \ref{fig:center_bias}. To execute spatial importance sampling, we map this frequency distribution into a bounded inverse weight scale:
 \begin{equation}
    \mathbf{W}_{spatial}(x,y) = w_{max} - (w_{max} - w_{min}) \cdot \tilde{F}(x,y)
\end{equation}
where $w_{max}$ and $w_{min}$ define the empirical upper and lower bounds. In our implementation, we set $w_{min} = 0.2$ and $w_{max} = 3.0$. Serving as explicit spatial importance sampling, this inverse scaling mapping heavily penalizes loss on long-tail samples, thereby fundamentally suppressing the inherent center-bias.

The global gaze map prediction is a comprehensive task that requires learning both the overall spatial distribution and the precise pixel-wise numerical values. Therefore, the global branch is supervised by combining KL loss and spatial-weighted BCE loss:
\begin{equation}
    \mathcal{L}_{global} = \mathcal{L}_{wbce} + \alpha \mathcal{L}_{kld}(\mathbf{G}_{global}, \hat{\mathbf{M}}_{global})
\end{equation}
where $\alpha$ is a scaling factor balancing the two constraints.

\subsubsection{Total Objective Function}
The total optimization objective integrates these constraints into a unified weighted sum:
\begin{equation}
    \mathcal{L}_{total} = \mathcal{L}_{text} + \lambda_{1} \mathcal{L}_{region} + \lambda_{2} \mathcal{L}_{global}
\end{equation}
where $\lambda_{1}$ and $\lambda_{2}$ are empirical hyperparameters balancing linguistic reasoning, object-level distribution alignment, and global-level spatial alignment. 

\section{Experiments}

\subsection{Experimental Setup}
\subsubsection{Implementation Details}
All experiments are conducted on a workstation equipped with four NVIDIA RTX 4090 GPUs. The backbone VLM is Qwen3-VL-8B-Instruct, where the vision encoder is kept frozen and only the language model is fine-tuned via LoRA with rank $r{=}16$, $\alpha{=}32$, and dropout of 0.1. We employ the AdamW optimizer with a unified learning rate of $1 \times 10^{-4}$ and weight decay of 0.01, scheduled by cosine annealing with a 2\% linear warmup. The per-GPU micro-batch size is set to 2 with a gradient accumulation of 10 steps, yielding an effective batch size of 80 across 4 GPUs. The model is trained for 12 epochs on the full G-W3DA training set. Input images are processed through the native Qwen3-VL processor with $\text{min\_pixels}{=}200{,}704$ and $\text{max\_pixels}{=}1{,}003{,}520$,
and gaze maps are resized to $256 \times 256$. We employ a mixed loss function with $\lambda_1=0.2, \lambda_2=3.0, \alpha=0.5$.

\subsubsection{Evaluation Metrics}
To comprehensively evaluate the spatial alignment and distribution similarity of the predicted gaze maps against the human ground truth, we adopt five standard saliency evaluation metrics. These metrics analyze the predictions from both distribution-based and location-based perspectives:

\begin{enumerate}
    \item \textbf{Kullback-Leibler (KL) Divergence:} Measures the relative entropy between the predicted probability density $P$ and the ground truth distribution $G$. A lower KL score indicates better structural alignment.
    \item \textbf{Pearson's Correlation Coefficient (CC):} Computes the linear correlation between the predicted and ground truth maps. It is formulated as:
    \begin{equation}
        \text{CC}(P, G) = \frac{\sigma(P, G)}{\sigma(P)\sigma(G)}
    \end{equation}
    where $\sigma(P, G)$ is the covariance, and $\sigma(P), \sigma(G)$ are standard deviations.
    \item \textbf{Similarity (SIM):} Measures the intersection of two normalized distributions:
    \begin{equation}
        \text{SIM}(P, G) = \sum_{x,y} \min(P(x,y), G(x,y))
    \end{equation}
    \item \textbf{Area Under Curve (AUC):} Treats gaze estimation as a binary classification task and plots the Receiver Operating Characteristic (ROC) curve. We employ both the Judd (AUC-J) \cite{AUC-J} and Borji (AUC-B) \cite{AUC-B} variants for evaluation.
\end{enumerate}

\begin{figure}[t]
    \centering
    \includegraphics[width=\linewidth]{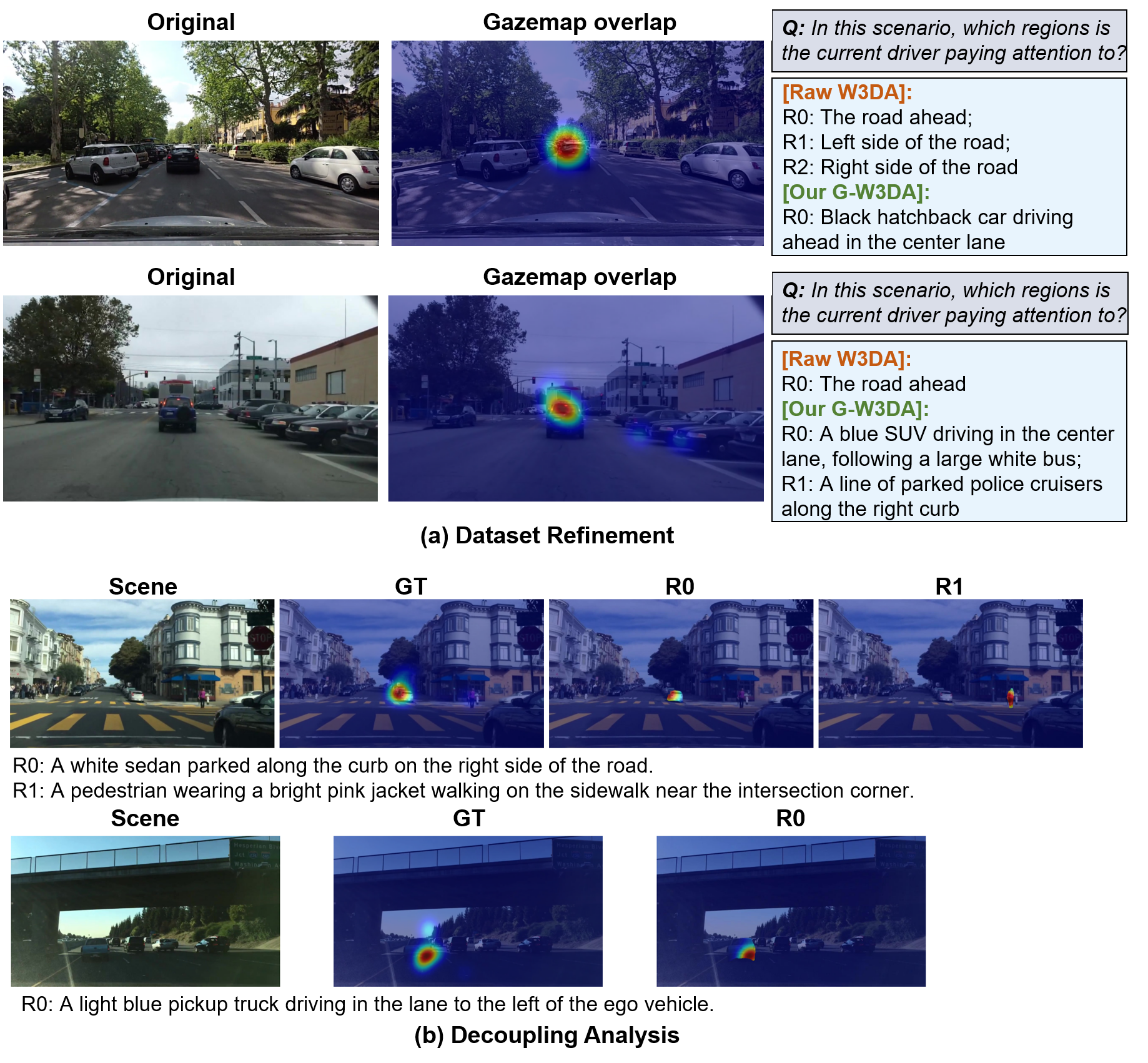}
    \caption{\textbf{Qualitative evaluation of dataset refinement and attention decoupling.} 
    (a) \textbf{Dataset Refinement:} Two comparative cases demonstrating the precision of our semantic parsing. The top case shows G-W3DA precisely anchoring attention to the specific ``black hatchback car'' missed by raw ambiguous annotations. The bottom case illustrates the successful recovery of secondary cognitive targets with weaker attention density via our hierarchical pipeline. 
    (b) \textbf{Decoupling Analysis:} Our refinement pipeline converts diffuse scene-level responses into sharply localized object-level masks, spatially aligning the risk distribution with actual semantic targets.}
    \label{fig:dataset_refinement}
\end{figure}

\begin{figure}[t]
    \centering
    \includegraphics[width=0.98\linewidth]{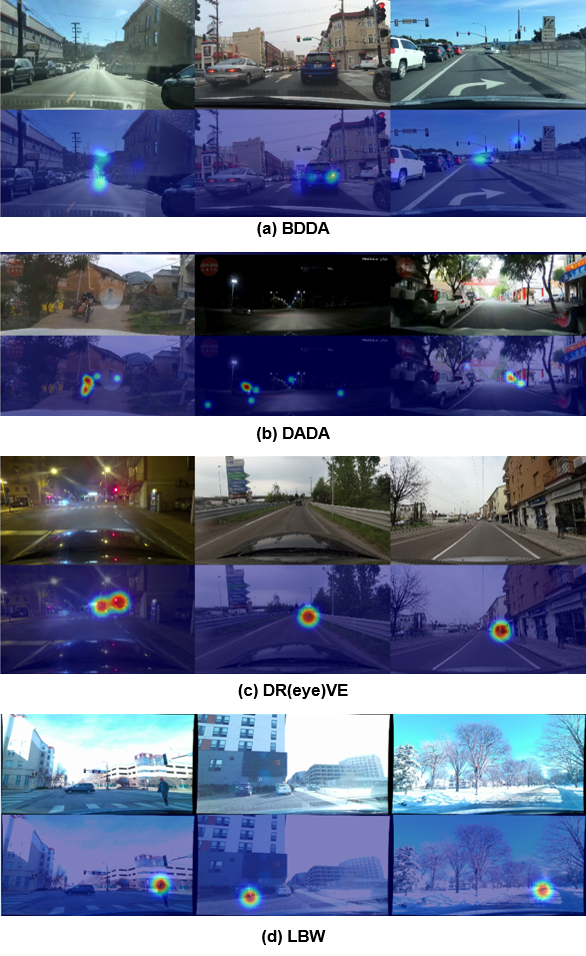} 
    \caption{\textbf{Diversity of driving scenarios in the W3DA benchmark.} Within each dataset block, the top row displays the original raw driving scenes, while the bottom row illustrates the corresponding ground-truth attention heatmaps overlaid on the images. The benchmark encompasses diverse conditions: (a) BDDA (safety-critical), (b) DADA (accident scenarios), as well as (c) DR(eye)VE and (d) LBW (normal driving).}
    \label{fig:w3da_samples}
\end{figure}

\subsection{Dataset Refinement Analysis}
Before evaluating the model's predictive performance, we analyze the quality and diversity of the newly constructed G-W3DA dataset. Fig. \ref{fig:w3da_samples} illustrates sample scenarios from the four distinct sub-datasets. The dataset encompasses an extensive variety of conditions, featuring extreme variability in weather (e.g., sunny, snowy), lighting (e.g., daytime, nighttime), and scene types (e.g., urban intersections, highways). Furthermore, it comprehensively covers a wide spectrum of driving situations, ranging from safety-critical near-crashes (BDDA) and actual traffic accidents (DADA) to normal driving tasks (DR(eye)VE and LBW).

We further evaluate the granularity of our annotations through qualitative analysis, as visualized in Fig. \ref{fig:dataset_refinement}. Fig. \ref{fig:dataset_refinement}(a) provides two comparative cases demonstrating the precision of our semantic parsing. In the first case (top), our G-W3DA precisely anchors attention to the specific ``black hatchback car'', which is missed by the raw ambiguous annotations. The second case (bottom) illustrates the advantage of our hierarchical cross-validation protocol: it successfully recovers secondary cognitive targets with weaker attention density, such as the parked vehicles on the curb. Furthermore, the qualitative comparison in Fig. \ref{fig:dataset_refinement}(b) shows that our refinement pipeline converts scene-level responses into sharply localized object-level masks, spatially aligning the risk distribution with the actual semantic targets rather than the surrounding background. This semantic sharpening effectively enforces strict alignment between textual descriptions and physical instance boundaries.

Table \ref{tab:dataset_quality} further summarizes the annotation quality across four major sub-datasets, comprising a total of 61,656 valid samples. A critical metric in this evaluation is the Mean Attention Intensity ($\mu_{attn}$). As formulated in Equation \ref{eq:mu_attn}, it is calculated as the average normalized human gaze probability within the extracted object-level region masks. Our G-W3DA brings a substantial enhancement in both the granularity and the cognitive fidelity of the data. Across all driving samples, the average number of identified risk regions per sample rises from 1.57 to 1.86, indicating more gazed targets. Interestingly, the extracted data reveals a strong positive correlation between the scene's risk level and the number of attended regions. In safety-critical (BDDA) and traffic accident (DADA) scenarios, the average regions per sample surge to 2.05 and 2.55, respectively, whereas normal driving subsets (DR(eye)VE and LBW) maintain a lower average of around 1.10. Notably, the quality of these spatial masks is superior. The mean attention intensity improves from 0.1791 to 0.2956, marking a remarkable 65.0\% relative improvement. In subsets like DR(eye)VE and LBW, while the number of regions slightly decreases, their mean attention intensity increases. This phenomenon validates the efficacy of our cross-validation mechanism, which filters out the hallucinated and task-irrelevant masks.

\begin{table}[t]
\centering
\caption{Comparison of dataset annotation quality. \textbf{Avg. Regions} denotes the average number of object masks extracted per scene. \textbf{Mean Attn Intensity} refers to the average normalized human gaze probability strictly within the boundaries of these masks.}
\label{tab:dataset_quality}
\resizebox{\columnwidth}{!}{
\begin{tabular}{l|c|cc|cc}
\toprule
\multirow{2}{*}{Dataset} & \multirow{2}{*}{Total Samples} & \multicolumn{2}{c|}{Avg. Regions per Sample} & \multicolumn{2}{c}{Mean Attn Intensity ($\mu_{attn}$)$\uparrow$} \\
\cmidrule{3-6}
& & Original & \textbf{G-W3DA (Ours)} & Original & \textbf{G-W3DA (Ours)} \\
\midrule
BDDA & 19,342 & 1.84 & 2.05 & 0.1460 & \textbf{0.2282} \\
DADA & 19,854 & 1.78 & 2.55 & 0.1373 & \textbf{0.2896} \\
DR(eye)VE & 18,244 & 1.17 & 1.10 & 0.2954 & \textbf{0.4389} \\
LBW & 4,216 & 1.01 & 1.00 & 0.2220 & \textbf{0.3512} \\
\midrule
\textbf{Overall} & \textbf{61,656} & 1.57 & 1.86 & 0.1791 & \textbf{0.2956} \\
\bottomrule
\end{tabular}
}
\end{table}

\subsection{Quantitative Comparison with State-of-the-Art}
We compared the proposed DualGaze-VLM with an extensive set of baselines on the W3DA benchmark, ranging from traditional CNN-based saliency models to the latest multimodal foundation models (LLada \cite{11-zhu2025llada}, FSDAM \cite{14-hamid2025fsdam}). The quantitative results are comprehensively summarized in Table \ref{tab:sota_comparison}.

\subsubsection{Quantitative Comparison and Overall Superiority}
In the comprehensive evaluation across three major driving scenarios, our proposed DualGaze-VLM demonstrates exceptional overall performance, achieving state-of-the-art (SOTA) results and comprehensively surpassing existing baselines in spatial alignment metrics, particularly under Safety-Critical situations. Notably, on the SIM and CC metrics---which measure the internal topological consistency of probability distributions---our method achieves significant improvements of approximately 17.8\% (from 0.467 to 0.550) and 4.9\% (from 0.589 to 0.618), respectively, compared to the previous best model FSDAM. This substantial margin directly validates the effectiveness of our \textbf{text-guided object-level decoupling} mechanism: it not only localizes the general gaze area but perfectly reconstructs the highly focused physical boundaries inherent to human attention when facing hazards. Furthermore, in the Normal Driving scenario, our model exhibits a marginal lag in metrics such as KLdiv (1.229 vs 1.219). Rather than a deficiency, this represents a benign trade-off introduced by our Spatial-Weighted BCE loss function. In mundane scenarios, traditional models tend to overfit the strong "center-bias" prior to obtain higher scores. By explicitly counteracting this prior, our model sacrifices a negligible amount of central fitting accuracy to ensure that, in emergencies, attention can be acutely shifted to peripheral risk targets, successfully avoiding the "blindly centered" visual hallucination.

\subsubsection{Synergy of Multi-task and Cognitive Prior Injection}
The experimental results profoundly reveal a paradigm shift, where multi-modal, multi-task collaborative architectures decisively outperform traditional pure-vision attention mechanisms. As shown in Table \ref{tab:sota_comparison}, powerful pure-vision deep networks like ConvNeXt and ERFNet achieve high AUC scores in simple scenarios but maintain stubbornly high KLdiv metrics (e.g., ConvNeXt at 1.765) in complex, Safety-Critical situations, falling far behind the 1.2 level of multi-modal baselines. This indicates that bottom-up visual saliency feature extraction has hit a performance ceiling in processing driving decisions. In contrast, multi-task models supervised by textual reasoning (such as LLada, FSDAM, and ours) demonstrate overwhelming advantages. DualGaze-VLM extracts the hidden states of task instructions via VLM as a unified semantic query vector, injecting top-down causal reasoning logic into visual feature modulation. This mechanism frees the model from relying solely on pixel contrast; instead, it genuinely comprehends "why to look" and "what to look at," thereby achieving high-precision visual anchoring in complex causal scenarios.

\subsubsection{Cross-Domain Stability and Robustness Analysis}
In the Traffic Accident subset, nearly all baseline models suffer from severe performance degradation. This phenomenon primarily stems from the inherent domain shift in this data split (e.g., derived from the DADA dataset's dashcam perspective), which contains substantial watermark and text interference, along with distinct optical distortions and camera heights compared to standard first-person driver views. Despite these formidable cross-domain challenges and visual noise, DualGaze-VLM maintains the most stable performance, registering the minimal performance drop (e.g., KL divergence remains at an excellent 1.700, and CC sustains at 0.471). This remarkable robustness is primarily attributed to the deep synergy between our prior object-level data construction paradigm and the network architecture. During training, the G-W3DA dataset provides the model with denoised, boundary-pure object-level supervision. Guided by this high-quality prior, the dual-branch network leverages the unified semantic query vector to learn robust, "intent-driven" visual representations, rather than merely fitting fragile low-level pixel patterns. Consequently, when confronted with an unknown and noisy visual domain, the model relies on its strong semantic cognitive representations to effectively resist watermark and perspective interference, accurately mapping the core intent of the driver.

\begin{table*}[t]
\centering
\caption{Comparison of attention map prediction performance. \textbf{Bold} and \underline{underline} indicate the best and second-best results. $\dagger$ marks multi-task models jointly optimizing Attention Map Prediction and Textual Explanation Generation.}
\label{tab:sota_comparison}
\resizebox{\textwidth}{!}{
\begin{tabular}{l|ccccc|ccccc|ccccc}
\toprule
\multirow{2}{*}{Method} & \multicolumn{5}{c|}{Normal Driving} & \multicolumn{5}{c|}{Safety-Critical Situation} & \multicolumn{5}{c}{Traffic Accident} \\
\cmidrule{2-16}
& KLdiv$\downarrow$ & CC$\uparrow$ & SIM$\uparrow$ & AUC\_J$\uparrow$ & AUC\_B$\uparrow$ & KLdiv$\downarrow$ & CC$\uparrow$ & SIM$\uparrow$ & AUC\_J$\uparrow$ & AUC\_B$\uparrow$ & KLdiv$\downarrow$ & CC$\uparrow$ & SIM$\uparrow$ & AUC\_J$\uparrow$ & AUC\_B$\uparrow$ \\
\midrule
ITTI~\cite{ITTI} & 3.216 & 0.093 & 0.080 & 0.676 & 0.665 & 2.807 & 0.049 & 0.119 & 0.618 & 0.613 & 3.339 & 0.033 & 0.073 & 0.595 & 0.600 \\
GBVS~\cite{GBVS} & 2.572 & 0.294 & 0.139 & 0.868 & 0.857 & 2.238 & 0.246 & 0.176 & 0.839 & 0.804 & 2.826 & 0.173 & 0.105 & 0.814 & 0.765 \\
DeepGaze I~\cite{DeepGazeI} & 3.103 & 0.207 & 0.080 & 0.885 & 0.546 & 2.550 & 0.237 & 0.137 & 0.874 & 0.595 & 3.060 & 0.206 & 0.082 & 0.883 & 0.673 \\
DeepGaze IIE~\cite{DeepGazeIIE} & 3.071 & 0.226 & 0.082 & 0.887 & 0.610 & 2.505 & 0.296 & 0.141 & 0.926 & 0.589 & 3.026 & 0.227 & 0.085 & 0.917 & 0.584 \\
MLNet~\cite{MLNet} & 2.129 & 0.547 & 0.460 & 0.914 & 0.836 & 1.953 & 0.528 & 0.433 & 0.928 & 0.874 & 2.897 & 0.344 & 0.288 & 0.893 & 0.784 \\
CDNN~\cite{CDNN} & 2.614 & 0.465 & 0.394 & 0.887 & 0.790 & 2.646 & 0.401 & 0.350 & 0.885 & 0.767 & 3.568 & 0.283 & 0.254 & 0.851 & 0.714 \\
FBnet~\cite{FBNet} & 2.980 & 0.406 & 0.343 & 0.869 & 0.769 & 2.585 & 0.431 & 0.364 & 0.887 & 0.803 & 3.197 & 0.329 & 0.246 & 0.873 & 0.789 \\
ConvNeXt~\cite{ConvNext} & 2.042 & 0.570 & 0.412 & 0.916 & 0.848 & 1.765 & 0.567 & 0.413 & 0.938 & 0.877 & 3.049 & 0.377 & 0.248 & 0.891 & 0.806 \\
ERFNet~\cite{ERFNet} & 1.979 & 0.558 & 0.425 & 0.923 & 0.840 & 1.593 & 0.538 & 0.410 & 0.942 & 0.868 & 2.181 & 0.391 & 0.253 & 0.930 & 0.846 \\
GazeXplain$\dagger$~\cite{GazeXPlain} & 2.578 & 0.477 & 0.389 & 0.857 & 0.866 & 2.769 & 0.383 & 0.321 & 0.848 & 0.743 & 3.109 & 0.371 & 0.236 & 0.902 & 0.804 \\
LLada$\dagger$~\cite{LLaDa} & \textbf{1.219} & \underline{0.583} & 0.436 & 0.952 & \textbf{0.908} & 1.230 & 0.579 & 0.420 & 0.950 & \underline{0.912} & 1.927 & 0.396 & 0.262 & 0.934 & \underline{0.889} \\
FSDAM$\dagger$~\cite{FSDAM} & 1.408 & 0.577 & \underline{0.467} & \underline{0.959} & 0.862 & \underline{1.222} & \underline{0.589} & \underline{0.467} & \underline{0.958} & 0.890 & \underline{1.838} & \underline{0.439} & \underline{0.323} & \underline{0.949} & 0.872 \\
\midrule
\textbf{DualGaze-VLM (Ours) $\dagger$} & \underline{1.229} & \textbf{0.613} & \textbf{0.468} & \textbf{0.964} & \underline{0.900} & \textbf{1.106} & \textbf{0.618} & \textbf{0.550} & \textbf{0.961} & \textbf{0.918} & \textbf{1.700} & \textbf{0.471} & \textbf{0.325} & \textbf{0.952} & \textbf{0.898} \\
\bottomrule
\end{tabular}
}
\end{table*}

\begin{figure*}[t]
    \centering
    \includegraphics[width=0.8\textwidth, keepaspectratio]{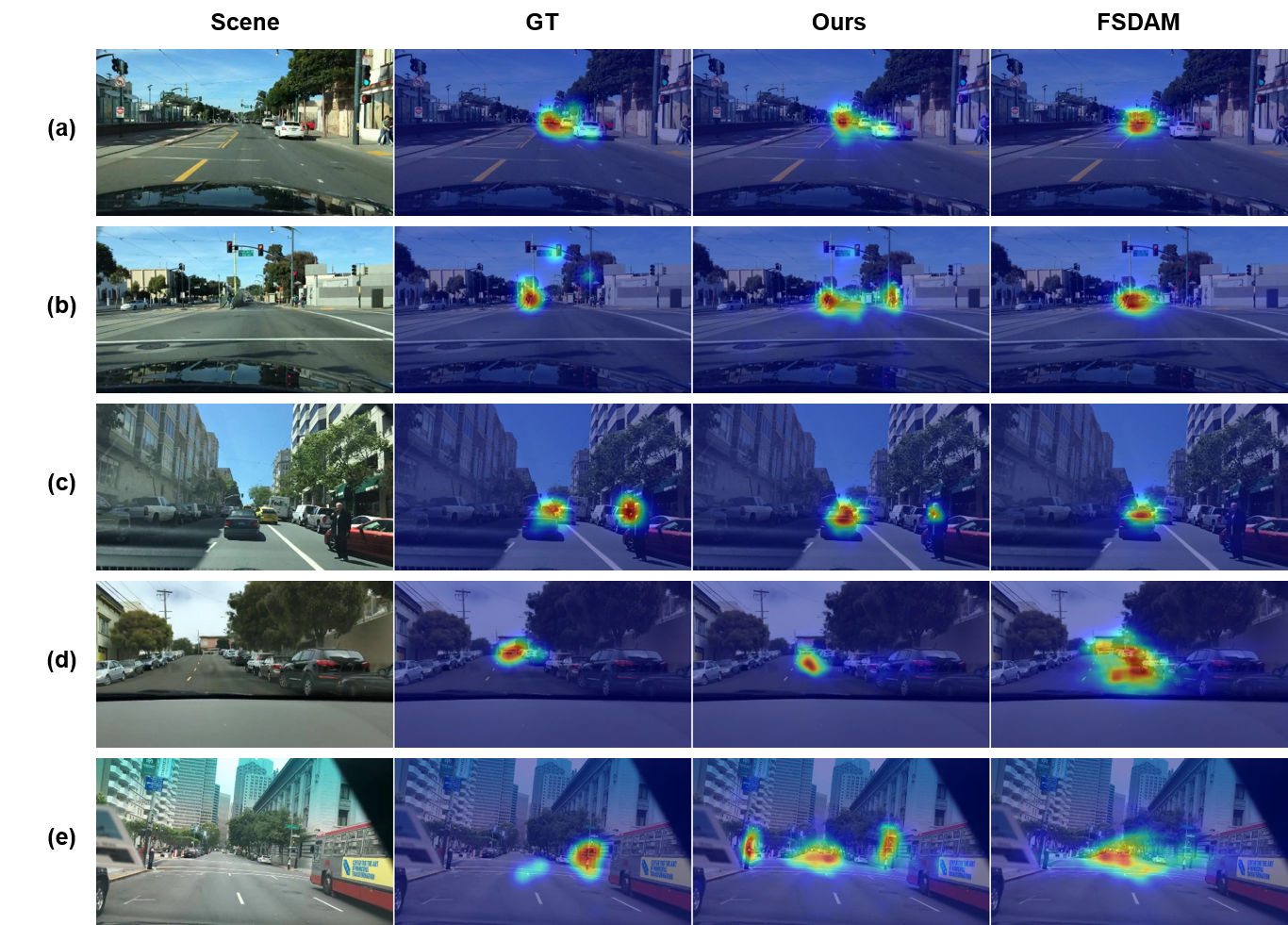}
    \caption{\textbf{Qualitative comparison of global attention prediction.} Our method generates attention maps that are more consistent with ground truth than FSDAM, with better coverage of multiple hazard-relevant regions, stronger sensitivity to pedestrians and other vulnerable road users, and more compact responses with less diffuse activation. It also clearly mitigates the center-bias of the baseline by shifting attention toward the truly relevant peripheral hazards.}
    \label{fig:qualitative_results}
\end{figure*}

\subsection{Qualitative Analysis and Visualization}
As shown in Fig. \ref{fig:qualitative_results}, our method produces attention maps that are more consistent with the ground-truth distribution than FSDAM. In particular, it is more responsive to multiple critical regions in the scene, more sensitive to vulnerable road users such as pedestrians, and less prone to diffuse low-intensity activation over broad areas. In addition, the qualitative results reveal that FSDAM suffers from a noticeable center-bias, i.e., a tendency to place excessive probability mass near the image center even when the true hazard lies away from it. Benefiting from the explicit reasoning sequence, our global branch effectively alleviates this issue and shifts attention toward the actual hazard-relevant regions, especially in peripheral areas.

Specifically, \textbf{Rows~(a) and~(b)} illustrate multi-target scenarios, where our method better captures the distributed attention pattern and covers multiple relevant regions more completely. \textbf{Row~(c)} shows a pedestrian-related urban driving case, in which our method is more sensitive to the pedestrian while still preserving attention to the forward vehicle. \textbf{Rows~(d) and~(e)} further demonstrate that our method yields more compact and semantically focused attention maps, avoiding unnecessary diffuse responses. Notably, all five rows also reflect the advantage of reduced center-bias, as our method more effectively shifts attention away from the image center and toward the truly relevant hazard regions.

\subsection{Ablation Study}
To rigorously validate the effectiveness of our proposed contributions, we conduct a comprehensive ablation study focusing on two core dimensions: the data paradigm and the architectural design. To ensure a fair comparison, we applied the exact same SAM-based annotation pipeline described in Section \ref{s:pipeline} to extract object-level region masks for the W3DA dataset based on its raw descriptions. All models in the ablation study were uniformly trained for 12 epochs under identical hyperparameter settings. In addition to the global saliency metrics, we also evaluated the object-level region saliency which is extracted by applying region masks to the global gaze maps. The quantitative results of these configurations are detailed in Table \ref{tab:ablation_study}.

\subsubsection{Visual-Bias Hallucination and SE-Gate Degeneration}
Observing the Object-Level Region Saliency results in Table \ref{tab:ablation_study}, a critical anomaly emerges: even when the SAM cross-validation method is identically applied to the original dataset, integrating the SE-Gate yields no significant improvement, and even results in a marginal degradation in KLdiv (from 2.368 to 2.369). The root cause of this failure lies not in the architectural design, but in the severe \textbf{Visual-Bias Hallucination} inherent in traditional dataset annotation paradigms. Because the original dataset annotation relies on joint-input prompting (feeding RGB and global heatmaps simultaneously), the VLM frequently hallucinates based on pre-trained visual priors, generating semantic descriptions that are fundamentally misaligned with the actual driver gaze distribution. Feeding these hallucinated, misaligned texts into SAM inevitably yields highly noisy and inaccurate region masks. When the SE-Gate receives semantic instructions that are severely misaligned with the spatial supervision masks (a classic "garbage in, garbage out" dilemma), the network encounters severe gradient contradiction and fails to establish valid text-vision correspondences, rendering the gating mechanism ineffective.

\subsubsection{The Synergy of Decoupled Data and Condition-Aware Routing}
In stark contrast to the failure on the original dataset, the combination of our G-W3DA dataset and the SE-Gate triggers a profound synergy effect. As shown in Table \ref{tab:ablation_study}, utilizing the G-W3DA dataset alone without the SE-Gate only marginally improves the Region SIM to 0.217. Similarly, using the SE-Gate on the original dataset yields a Region SIM of 0.217. However, when both components are integrated, the Region SIM surges to 0.239 (an over 10\% relative improvement), and the KLdiv drops significantly to 2.280. This breakthrough demonstrates that architectural innovations alone are insufficient to bridge the modality gap. The rigorously decoupled, hallucination-free physical boundaries provided by G-W3DA act as a pristine "target" for the model, while the Condition-Aware SE-Gate provides the "dynamic routing" capability. Both components are indispensable prerequisites for mastering precise text-object cognitive grounding.

\begin{table}[t]
\centering
\caption{Ablation study on the effectiveness of the Condition-Aware SE-Gate and the Restructured G-W3DA Dataset. Performance is comprehensively evaluated on the Object-Level Region Saliency task.}
\label{tab:ablation_study}
\resizebox{\columnwidth}{!}{
\begin{tabular}{cc|ccccc}
\toprule
\multirow{2}{*}{Dataset} & \multirow{2}{*}{SE-Gate} & \multicolumn{5}{c}{Object-Level Region Saliency} \\
\cmidrule{3-7}
& & KLdiv$\downarrow$ & CC$\uparrow$ & SIM$\uparrow$ & AUC\_J$\uparrow$ & AUC\_B$\uparrow$ \\
\midrule
W3DA & $\times$  & 2.368 & 0.404 & 0.212 & 0.954 & 0.901 \\
W3DA & \checkmark & 2.369 & 0.403 & 0.217 & 0.955 & 0.897 \\
G-W3DA & $\times$ & 2.325 & 0.413 & 0.217 & 0.957 & \textbf{0.903} \\
\midrule
G-W3DA & \checkmark & \textbf{2.280} & \textbf{0.417} & \textbf{0.239} & \textbf{0.959} & 0.889 \\
\bottomrule
\end{tabular}
}
\end{table}
\subsection{Human Evaluation Test}
While quantitative metrics such as KL divergence provide objective mathematical comparisons, they do not entirely capture the physiological naturalness and cognitive plausibility of the generated attention maps. To rigorously assess whether our generated heatmaps are perceptually indistinguishable from actual human visual focus, we conduct a subjective human evaluation inspired by the Turing Test \cite{Turing2009}.

The evaluation dataset consists of 120 samples: 60 ground-truth human attention heatmaps randomly sampled from the dataset, and 60 corresponding heatmaps generated by our proposed DualGaze-VLM. A total of 20 subjects with varying driving experience (including diverse driving years and annual mileages) participate, yielding 2,400 evaluated samples. All participants provided informed consent prior to their involvement in the study.

During the evaluation, subjects view a dual-panel interface displaying the raw driving scene alongside the heatmap overlay. We explicitly instruct subjects to judge whether the spatial attention distribution aligns with the cognitive logic of a real human driver. For each sample, subjects make a binary decision regarding the source of the gaze map.

\begin{figure}[t]
    \centering
    \includegraphics[width=0.95\linewidth]{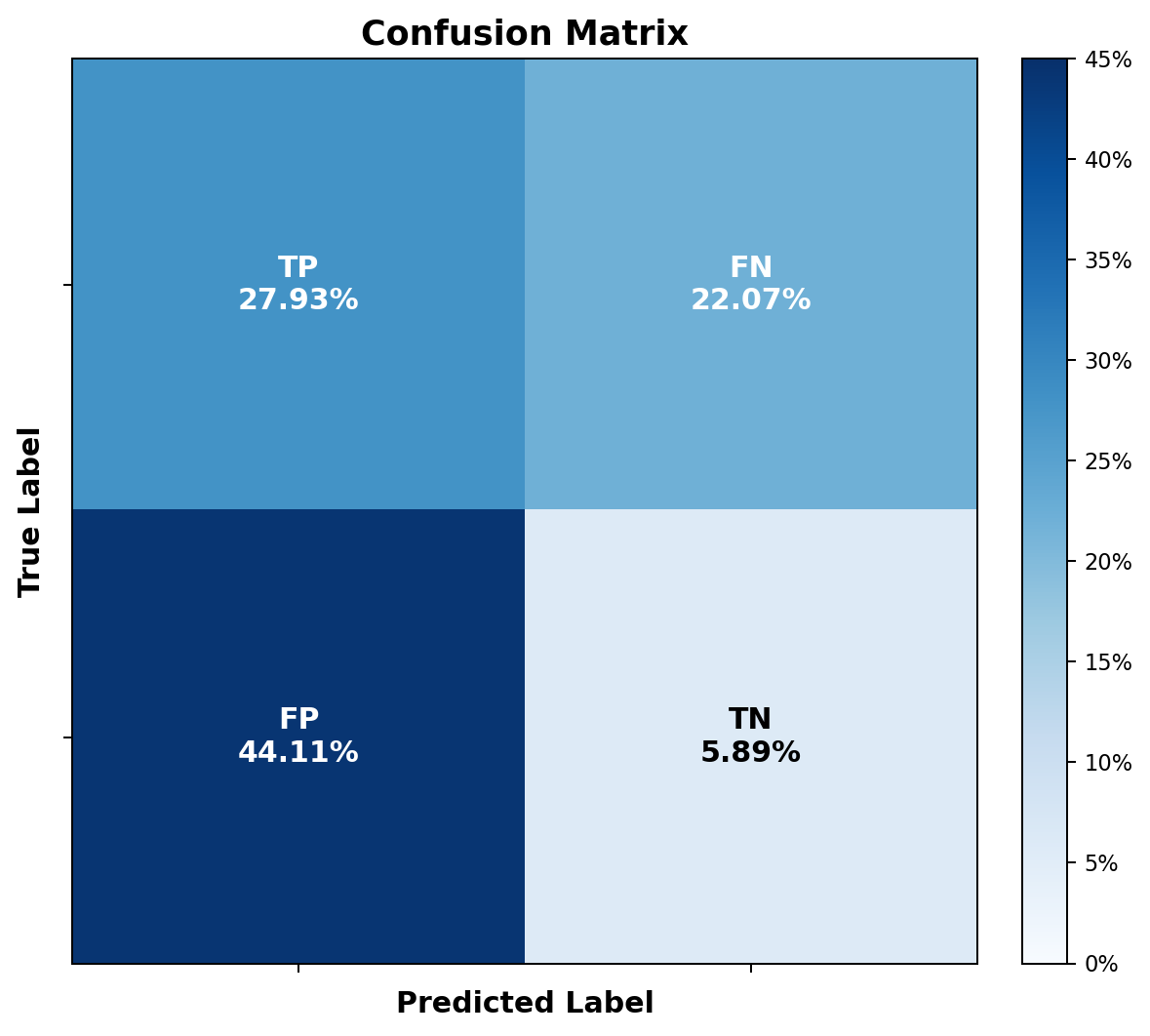}
    \caption{Confusion Matrix of the human evaluation test. The matrix illustrates the subjective classification results of 2,400 samples by 20 human evaluators. True labels represent the actual source (Real Human vs. DualGaze-VLM), while Predicted labels represent the subjects' perceptual judgments.}
    \label{fig:confusion_matrix}
\end{figure}

The evaluation results are quantitatively visualized in the confusion matrix in Figure \ref{fig:confusion_matrix}. When evaluating the actual ground-truth heatmaps (Top Row), subjects correctly identify 55.86\% of them as real human attention. This accuracy is slightly above random choice, indicating that human evaluators possess a basic, reliable intuition for visual attention. The most interesting finding emerges from the evaluation of our model's generated heatmaps (Bottom Row). 88.22\% of the heatmaps generated by DualGaze-VLM are perceived as authentic human gaze by the evaluators, while only 11.78\% are correctly recognized as synthetic. These human evaluation results demonstrate that DualGaze-VLM generates highly realistic attention distributions, appearing even more rational and visually plausible to human observers than the raw ground-truth data itself.

\section{Conclusion}
In this paper, we address the critical modality gap bottleneck of VLMs in autonomous driving by proposing a novel text-guided object-level gaze prediction paradigm. We first identify the severe "visual-bias hallucination" issues inherent in traditional holistic attention datasets and innovatively construct G-W3DA, a large-scale object-level dataset. Through rigorous cross-modal cross-validation and spatial decoupling, this dataset provides hallucination-free, pure object-level physical boundary supervision for cognitive models. Building upon this robust data foundation, our proposed DualGaze-VLM framework successfully breaks the traditional reliance on static visual features. By introducing a Condition-Aware SE-Gate, the model dynamically routes multi-scale visual features utilizing causal semantic instructions extracted from the VLM, truly empowering the attention prediction process with the profound textual reasoning capabilities of foundation models. Extensive quantitative evaluations and qualitative analyses demonstrate that our method not only achieves SOTA performance across all key saliency distribution metrics with remarkable cross-domain robustness, but also precisely generates fine-grained gaze maps strictly aligned with natural language instructions. Future work will focus on optimizing the precision of object-level attention prediction and exploring the direct integration of these fine-grained spatial priors into the end-to-end autonomous driving systems. The G-W3DA dataset will be continuously updated at \url{https://cloud.tsinghua.edu.cn/d/0e9327ebe9c7483899bc}.
\bibliographystyle{IEEEtran}
\bibliography{main}
\end{document}